%% file: NeurIPS.tex
\def\year{2021}\relax
\documentclass[letterpaper]{article} % DO NOT CHANGE THIS
\usepackage{aaai21}  % DO NOT CHANGE THIS
\usepackage{times}  % DO NOT CHANGE THIS
\usepackage{helvet} % DO NOT CHANGE THIS
\usepackage{courier}  % DO NOT CHANGE THIS
\usepackage[hyphens]{url}  % DO NOT CHANGE THIS
\usepackage{graphicx} % DO NOT CHANGE THIS
\usepackage{natbib}  % DO NOT CHANGE THIS AND DO NOT ADD ANY OPTIONS TO IT
\usepackage{caption} % DO NOT CHANGE THIS AND DO NOT ADD ANY OPTIONS TO IT
\usepackage[utf8]{inputenc} % allow utf-8 input
\usepackage{nicefrac}       % compact symbols for 1/2, etc.
\usepackage{microtype}      % microtypography

\usepackage{booktabs}
\usepackage{amsmath,amssymb,amsfonts,pifont}
\usepackage{amsthm}
\usepackage{algorithmic}
\usepackage{multirow}
\usepackage{subfigure}
\usepackage{tabularx}
\usepackage{textcomp}
\usepackage{makecell}
\PassOptionsToPackage{table,xcdraw,dvipsnames}{xcolor}
\usepackage{colortbl}
\usepackage{soul}
\usepackage{bm}
\usepackage{xr}
\usepackage[switch]{lineno}
\usepackage{amsthm}
\usepackage{algorithmic}
\usepackage[linesnumbered,ruled,vlined]{algorithm2e}
\SetKwInput{KwInput}{Input}                % Set the Input
\SetKwInput{KwOutput}{Output}              % set the Output

\makeatletter
\newcommand*{\addFileDependency}[1]{% argument=file name and extension
  \typeout{(#1)}
  \@addtofilelist{#1}
  \IfFileExists{#1}{}{\typeout{No file #1.}}
}
\makeatother

\usepackage{xcolor}
\usepackage{mathtools}

\newcommand{\bx}{\bm{x}}%
\newcommand{\by}{\bm{y}}%
\newcommand{\bz}{\bm{z}}%
\newcommand{\bxit}{\bm{x}_{it}}

\newcommand{\Xstar}{X_{*}}
\newcommand{\bmu}{\bm{\mu}}
\newcommand{\bmuv}{\bm{\mu}^{(v)}}
\newcommand{\bmui}{\bm{\mu}^{(i)}}
\newcommand{\kgamma}{k^{(v)}_{\gamma}}
\newcommand{\kphi}{k^{(i)}_{\phi}}
\newcommand{\ktheta}{k_{\theta}}
\newcommand{\tf}{\textbf{\text{f}}}
\newcommand{\tu}{\textbf{\text{u}}}
\newcommand{\tI}{\text{I}}
\newcommand{\norm}[1]{\left\lvert#1\right\rvert}
\newcommand{\Norm}[1]{\left\lVert#1\right\rVert}
\newcommand{\bep}{\bm{\epsilon}}

\newcommand{\para}[1]{\vspace*{1ex}\noindent\textbf{#1} }

\newcommand{\ie}{\mbox{\it{i.e.,\ }}}
\newcommand{\eg}{\mbox{\it{e.g.,\ }}}

\title{Longitudinal Deep Kernel Gaussian Process Regression}
\author{
  Junjie Liang, 
  Yanting Wu, 
  Dongkuan Xu, 
  Vasant Honavar\\
}
\affiliations{
    Pennsylvania State University \\
    \{jul672, yxw514, dux19, vhonavar\}@psu.edu
}

\begin{document}
% \linenumbers

\maketitle

\begin{abstract}
Gaussian processes offer an attractive framework for predictive modeling from longitudinal data, \ie irregularly sampled, sparse observations from a set of individuals over time. However, such methods have two key shortcomings: (i) They rely on ad hoc heuristics or expensive trial and error to choose the effective kernels, and (ii) They fail to handle multilevel correlation structure in the data. We introduce Longitudinal deep kernel Gaussian process regression (L-DKGPR) to overcome these limitations by fully automating the discovery of complex multilevel correlation structure from longitudinal data.  Specifically, L-DKGPR eliminates the need for ad hoc heuristics or trial and error using a novel adaptation of deep kernel learning that combines the expressive power of deep neural networks with the flexibility of non-parametric kernel methods. L-DKGPR effectively learns the multilevel correlation with a novel additive kernel that simultaneously accommodates both time-varying and the time-invariant effects. We derive an efficient algorithm to train L-DKGPR using latent space inducing points and variational inference. Results of extensive experiments on several benchmark data sets demonstrate that L-DKGPR significantly outperforms the state-of-the-art longitudinal data analysis (LDA) methods.
\end{abstract}

\input{intro}
\input{prelim}
\input{method}
\input{experiment}

\input{conclusion}
\bibliography{NeurIPS}

% \clearpage
% \input{appendix}

\end{document}

%% file: intro.tex
\section{Introduction}
Longitudinal studies, which involve repeated observations, taken at irregularly spaced time points, for a set of individuals over time, are ubiquitous in many applications, e.g., in health, cognitive, social, and economic sciences.  Such studies are used to identify the time-varying as well as the time-invariant  factors associated with a particular outcome  of interest, \eg health risk \cite{hedeker2006longitudinal}, urban computing \cite{tang2020joint,hsieh2020explainable}. 
%Such studies are often conducted on different locations and environments, including individuals with diverse background and characteristics. 
Longitudinal data typically exhibit longitudinal correlation (LC), \ie correlations among the repeated observations of a given individual over time; and cluster correlation (CC), \ie correlations among observations across individuals, \eg due to the characteristics that they share among themselves e.g., age, demographics factors; or both, \ie multilevel correlation (MC). In general, the  structure of MC can be complex and a priori unknown. Failure to adequately account for the structure of MC in predictive modeling from longitudinal data can lead to misleading statistical inferences \cite{gibbons1997random,liang2019lmlfm}. % As we shall see below, 
It can be non-trivial to choose a suitable correlation structure that reflects the correlations present in the data.  The relationships between the covariates and outcomes of interest can be highly complex and non-linear. 
Furthermore, modern applications often call for LDA methods that scale gracefully with increasing number of variables, the number of individuals, and the number of longitudinal observations per individual.

\subsection{Related Work}
\subsubsection{Conventional LDA Methods}
LDA methods have been extensively studied for decades \cite{hedeker2006longitudinal,verbeke2014analysis}. Conventional LDA methods fall into to two broad categories: (i) marginal models and (ii) conditional models. Marginal models rely on assumptions about the marginal association among the observed outcomes. The generalized estimating equations (GEE) \cite{liang1986longitudinal}, where a working correlation matrix is specified to model the marginal association among the observed outcomes, offer an example of marginal models. The parameters of marginal models are often shared by all individuals in the population, yielding {\em population-averaged} effects or {\em fixed} effects. Conditional models on the other hand avoid directly specifying the full correlation matrix by distinguishing  {\em random} effects, \ie parameters that differ across individuals, from fixed effects, so as to estimate the individual parameters conditioned on the population parameters. A popular example of conditional models is the generalized linear mixed-effects models (GLMM) \cite{mcculloch1997maximum}.  
% Hence, the correlations of random effects across individuals specifies the correlations between their observed outcomes e.g., health status. 
Despite much work on both marginal and conditional models \cite{fitzmaurice2012applied,wang2014generalized,xiong2019mixed,liang2019lmlfm}, many of the challenges, especially the choice of correlation structure, and the selection of variables to model random versus fixed effects, and the scalability of the methods remain to be addressed.

\subsubsection{Non-parametric LDA Methods}
More recently, there is a growing interest in Gaussian processes (GP) \cite{quintana2016bayesian,cheng2019additive,wang2019exact} for LDA because of their advantages over conventional parametric LDA methods: (i) GP make fewer assumptions about the underlying data distribution by dispensing with the need to choose a particular parametric form of the nonlinear predictive model; (ii) GP permit the use of  parametertized kernels  to model the correlation between observed outcomes, to cope  with data sampled at irregularly spaced time points, by interpolating between samples; (iii) The interpretability of GP models can be enhanced by choosing modular kernels that are composed of simpler kernels that capture the shared correlation structure of a subset of covariates, and (iv) GP models can  flexibly account for both longitudinal and cluster correlations in the data. For example, 
% Quintana \etal \cite{quintana2016bayesian} define a Dirichlet Process prior on the kernel parameters, which enables a clustering structure across individuals; and employ a tri-diagonal auto-regressive kernel to approximate the longitudinal correlation,thereby substantially speeding up the computation of the inverse of the correlation matrix during model inference. 
\citet{cheng2019additive} utilize an additive kernel for Gaussian data and employ a step-wise search strategy to select the kernel components and covariates that optimize the predictive accuracy of the model. \citet{timonen2019interpretable} consider a heterogeneous kernel to model individual-specific (random) effects in the case of non-Gaussian data. Despite their advantages, existing  GP based approaches to LDA suffer from several shortcomings that limit their applicability in real-world settings: (i) The choice of an appropriate kernel often involves a tedious, often expensive and unreliable, process of trial and error \cite{rasmussen2003gaussian} or ad hoc heuristics for identifying a kernel or selecting a subset of kernels from a pool of candidates \cite{cheng2019additive}.  (ii) Suboptimal choice of kernels can fail to adequately model the complex MC structure in the data. (iii) They do not scale to thousands of covariates and/or millions of data points that are common in modern LDA applications. 

\subsection{Overview of contributions}
A key challenge in predictive modeling of longitudinal data has to do with modeling the complex correlation structure in the data.  We posit that the observed correlation structure is induced by the interactions between time-invariant, individual-specific effects, and time-varying population effects. Hence, we can divide the task of predictive modeling from longitudinal data into three sub-tasks: (i) Given an observed data set, how do we estimate the time-varying and time-invariant effects? (ii) Given the learned effects, how do we estimate the correlation structure present in the data? (iii) Given the correlation structure, how do we predict as yet unobserved, \eg future outcomes?

We introduce Longitudinal deep kernel Gaussian process regression (L-DKGPR) to fully automate the discovery of complex multi level correlation structure from longitudinal data.  L-DKGPR inherits the attractive features of GP while overcoming their key limitations. Specifically, L-DKGPR eliminates the need for ad hoc heuristics or trial and error by using a deep kernel learning method \cite{wilson2016deep} that combines the expressive power of deep neural networks with the flexibility of non-parametric kernel methods. L-DKGPR extends \cite{wilson2016deep} by introducing a novel additive kernel that includes two components, one for modeling the time-varying (fixed) effects and the other for modeling the time-invariant (random) effects, to compensate for the multilevel correlation structure in longitudinal data. To enhance the effectiveness and efficiency of model inference, we improve the inducing points technique by introducing inducing points directly in the latent space. Our formulation permits a tractable ELBO, which not only eliminates the need for Monte Carlo sampling, but also dramatically reduces the number of parameters and iterations needed to achieve state-of-the-art regression performance. 
% Our experiment results with both simulated and real-world benchmark longitudinal data show that L-DKGPR significantly outperforms, by large margins, the popular GLMM, GEE baselines as well as several state-of-the-art non-linear LDA methods \cite{liang2019lmlfm,timonen2019interpretable} and GP methods \cite{wilson2015kernel,salimbeni2018orthogonally}.

%% file: prelim.tex
\section{Preliminaries}
\label{sec:problem_definition}
\para{Notations.}
We denote a longitudinal data set by $\mathcal{D}=(X,\by)$, where $X\in\mathbb{R}^{N\times P}$ is the covariate matrix and $\by\in \mathbb{R}^{N\times 1}$ is the vector of measured outcomes. We denote a row in $X$ by $\bxit$, with $i,t$ indexing the individual and the time for the observation respectively. Because the observations for each individual are irregularly sampled over time, we have for each individual $i$, a submatrix $X_i\in\mathbb{R}^{N_i\times P} \subset X$, where $N_i$ is the number of observations available for the individual $i$. If we denote by $I$ be the number of individuals in $\mathcal{D}$, 
the covariate matrix $X$ is given by $X^\top=(X_1^\top,\cdots,X_I^\top)^\top$. Accordingly, the outcomes $\by$ are given by $\by^\top=(\by_1^\top,\cdots,\by_I^\top)^\top$. 

\para{Gaussian Process.}
A Gaussian process (GP) is a stochastic process, i.e., a distribution over functions or an infinite collection of (real-valued) random variables, such that any finite subset of random variables has a multivariate Gaussian distribution \cite{williams2006gaussian}. 
% The distribution of a Gaussian process is the joint distribution of all those (infinitely many) random variables, and hence a distribution over functions with a continuous domain, e.g. time or space. 
A kernel describes the covariance of the random variables that make up the GP. More precisely, if a function $f:\mathcal{X}\rightarrow \mathbb{R}$ has a GP prior $f {\sim} \mathcal{GP}(\mu,k_\gamma)$ where $\mu$ is the mean function and $k_{\gamma}(\cdot,\cdot)$ is a (positive semi-definite) kernel function parameterized by $\gamma$, then any finite collection of components of $f$ (denoted as $\tf$) has a multivariate Gaussian distribution $(\tf|X) {\sim} \mathcal{N}(\mu(X),K_{XX})$, where $\mu(X)$ is the mean vector, and $(K_{XX})_{ij}=k_\gamma(\bx_i,\bx_j)$ is the covariance matrix. In the regression setting, the function $f$ is treated as an unobserved signal linked to the outcomes through a (typically Gaussian) likelihood function, such that $(y|\tf){\sim}\mathcal{N}(\tf,\sigma^2\tI)$.
%  Given $\Xstar$ as the covariate matrix for the test data points, the predictive distribution is computed as

% \resizebox{0.95\linewidth}{!}{
% \begin{minipage}{\linewidth}
% \begin{equation}
% \label{eq:predictive}
% \nonumber
% (\tf_*|\Xstar,X,y){\sim}\mathcal{N}(\underbrace{\mu(\Xstar) + K_{\Xstar X}[K_{XX}+\sigma^2\tI]^{-1}\left(\by - \mu(X)\right)}_{\text{predictive mean follows observations}},\underbrace{K_{\Xstar \Xstar} - K_{\Xstar X}[K_{XX}+\sigma^2\tI]^{-1}K_{\Xstar X}^\top}_{\text{predictive variance shrinks with more data}})
% \end{equation}
% \end{minipage}
% }

% where $K_{\Xstar X}$ is the covariance matrix between $\Xstar$ and $X$, which can be computed using the same kernel function $k_\gamma$. 

%Because the kernel function plays the same role as a covariance matrix in finite input data, a valid kernel function should be positive semidefinite. 
% The radial basis function (RBF)  is a popular choice as a kernel function \cite{williams2006gaussian}. For a given pair of data points $\bx_i,\bx_j$, the RBF kernel is given by:
% \begin{equation}
%     k_{RBF}(\bx_i,\bx_j)=\exp\left(-\frac{1}{2}\sum_{k=1}^P \left(\frac{x_{ik}-x_{jk}}{l_k}\right)^2\right)
% \end{equation}
% with kernel parameters $\gamma=\{l_1,\cdots,l_P\}$. 
%To keep the notation simple, the RBF kernel function is abbreviated as $k$ throughout the rest of the paper unless stated otherwise.

{\em Additive  GP} is a special case of GP where unobserved signal is expressed as the sum of $J$ independent signal components, \ie $f=\sum_{j=1}^J\alpha_j f^{(j)}$, where $\bm{\alpha}=\{\alpha_j\}_{j=1}^J$ are the the coefficients associated with the individual components \cite{duvenaud2011additive}. In practice, each signal component is computed on a (typically small \cite{cheng2019additive,timonen2019interpretable}) subset of the observed covariates in $\bx$. The fact that each signal component has a GP prior ensures that the joint signal $f$ is also GP.
% such that Eq.~\eqref{eq:mvn} is rewritten as
% \begin{equation}
% (\tf|X) {\sim} \mathcal{N}(\sum_{j=1}^J\alpha_j \mu^{(j)}(X), \sum_{j=1}^J \alpha_j^2 k_{\gamma}^{(j)}(X,X))
% \end{equation}
Additive GP allows using different kernel functions for different signal components, so to model the shared correlation structure of a subset of covariates, thus enhancing the interpretability of the resulting GP. More importantly, it permits the time-varying and time-invariant effects to be modeled using different kernel functions, which is especially attractive in modeling longitudinal data.

% While it is possible to learn the kernel parameters of both GP and Additive GP using either MLE framework \cite{williams2006gaussian} or Bayesian approaches \cite{cheng2019additive,timonen2019interpretable}, both require  an iterative algorithm that requires inverting an $N$ by $N$ matrix at each iteration, yielding $\mathcal{O}(N^3)$ time complexity and $\mathcal{O}(N^2)$ space complexity per iteration in the training phase. The prohibitive computational complexity of the model limits the applicability of this approach in real-world settings with large numbers of observations. 
% Moreover, existing approaches for predictive modeling of longitudinal data suffer from lack of clear guidance for choosing an optimal set of kernels that are best suited to analyses of the data at hand. Hence, current methods for choosing kernels \cite{cheng2019additive,timonen2019interpretable} tend to be heuristic in nature, e.g., based on the data types of the covariates and is fixed throughout the subsequent analysis, which leads to sub-optimal correlation structure. 
%Such practice fails to capitalize the expressive power of GP, thus leading to a sub-optimal correlation structure. 

%% file: method.tex
\section{Longitudinal Deep Kernel Gaussian Process Regression}
\label{sec:londkgp}

% We start by defining the problem of predictive modeling from longitudinal data. Then we proceed to introduce L-DKGPR.

Predictive modeling from longitudinal data typically requires solving two sub-problems: (i) Extracting the time-varying and time-invariant information from the observed data to estimate the underlying multilevel correlation structure; and (ii) using the estimated correlation structure to predict the future outcomes. In what follows, we describe our solutions to both sub-problems.

\subsection{Modeling the Multilevel Correlation using Deep Kernels}
Recall that longitudinal data exhibit complex correlations  arising from the interaction between time-varying effects and time-invariant effects. Hence, we decompose the signal function $f$ into two parts, \ie $f^{(v)}$ which models the time-varying effects and $f^{(i)}$, which models the time-invariant effects. The result is a probabilistic model that can be specified as follows:
\begin{linenomath}
\begin{align}
\label{eq:gaussian_ll}
\nonumber
     & (\by|\tf) {\sim} \mathcal{N}(\tf,\sigma^2 \tI) \\\nonumber
     & f=\alpha^{(v)}f^{(v)} + \alpha^{(i)}f^{(i)}\\\nonumber
     & (\tf^{(v)}|X){\sim}\mathcal{N}(\bmuv(X),\kgamma(X,X)) \\\nonumber
     &(\tf^{(i)}|X){\sim}\mathcal{N}(\bmui(X),\kphi(X,X))
\end{align}
\end{linenomath}
% \resizebox{1\linewidth}{!}{
% \begin{minipage}{\linewidth}
% \begin{align*}
% \label{eq:gaussian_ll}
%     & (\by|\tf) {\sim} \mathcal{N}(\tf,\sigma^2 \tI), && f=\alpha^{(v)}f^{(v)} + \alpha^{(i)}f^{(i)}\\
%     & (\tf^{(v)}|X){\sim}\mathcal{N}(\bmuv(X),\kgamma(X,X)), &&(\tf^{(i)}|X){\sim}\mathcal{N}(\bmui(X),\kphi(X,X))
% \end{align*}
% \end{minipage}
% }
We denote the kernel parameters for time-varying effects and time-invariant effects respectively by $\gamma$ and $\phi$. The mean functions  $\bmuv,\bmui$, if unknown, can be estimated from data. In this study, without loss of generality, following \cite{williams2006gaussian,wilson2016deep,wilson2016stochastic,cheng2019additive,timonen2019interpretable}, we set $\bmuv=\bmui=0$. Assuming  that $\tf^{(v)}$ and $\tf^{(i)}$ are conditionally independent given $X$, we can express the joint signal distribution $\tf$ as follows:
\begin{linenomath}
\begin{equation}
\label{eq:overall_signal}
    (\tf|X){\sim}\mathcal{N}\left(\bm{0},\ktheta={\alpha^{(v)}}^2\kgamma+ {\alpha^{(i)}}^2\kphi\right)
\end{equation}
\end{linenomath}
%The structure of the time-varying kernel and time-invariant kernels are shown in Figs. \ref{fig:time_varying} and \ref{fig:time_invariant} respectively.

% \begin{figure}[t!]
% \vspace{-0.12cm}
% \centering 
% \subfigure[Time-varying kernel $\kgamma$]{
% \includegraphics[width=.35\linewidth]{fig/time_varying}
% \label{fig:time_varying} 
% }
% \subfigure[Time-invariant kernel $\kphi$]{
% \includegraphics[width=.35\linewidth]{fig/time_invariant}
% \label{fig:time_invariant}
% }
% \caption{Structure of the time-varying and time-invariant kernels. \fixme{change RBF to SE, squeeze the figures.}}
% \label{fig:kernel}
% \vspace{-0.1cm}
% \end{figure}

\begin{figure}[t!]
% \vspace{-0.12cm}
\centering 
\includegraphics[width=1\linewidth]{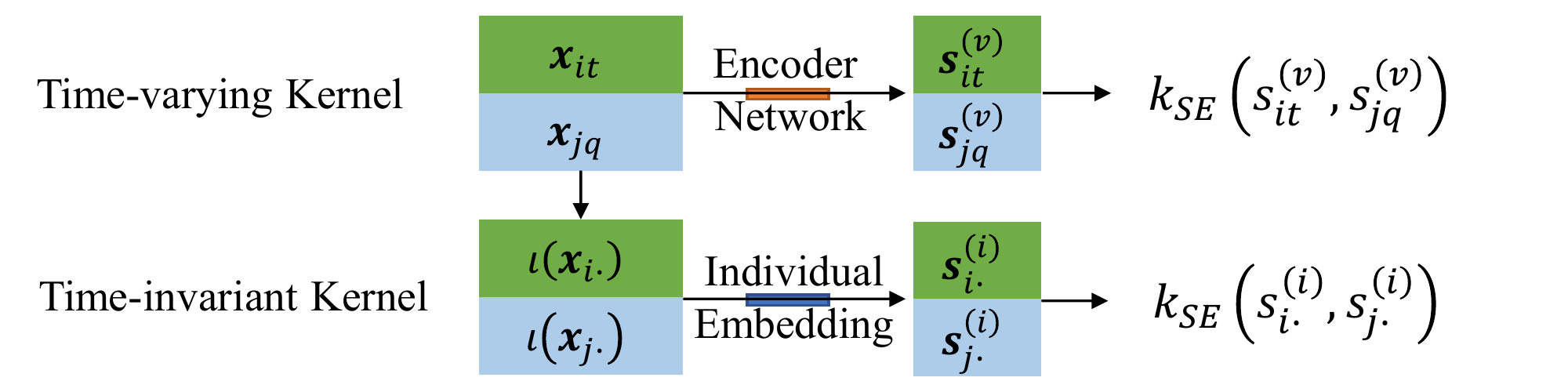}
% \vspace{-0.2cm}
\caption{Structure of the deep kernels.}
\label{fig:kernel}
\end{figure}
\para{Time-varying Kernel $k^{(v)}_{\gamma}$.}
We introduce a time-varying kernel to capture the longitudinal correlation in the data. The structure of our time-varying kernel $\kgamma$ is shown in the upper part of Figure~\ref{fig:kernel}. Let $e_{\gamma}:\mathcal{X}\rightarrow \mathcal{S}^{(v)}\in\mathbb{R}^{D_v}$ be a non-linear encoder function given by a deep architecture parameterized by $\gamma$. Given a pair of data points $\bx_{it},\bx_{jq}$, where $i,j$ index the individuals and $t,q$ index the time-dependent observations, the time-varying kernel is given by:
\begin{linenomath}
\begin{equation}
    \kgamma(\bx_{it},\bx_{jq}) = k_{SE}(e_{\gamma}(\bx_{it}),e_{\gamma}(\bx_{jq}))
\end{equation} 
\end{linenomath}
with $k_{SE}$ denoting the squared exponential kernel \cite{williams2006gaussian}. Note that SE kernel is based on Euclidean distance, which is not a useful measure of distance in the high dimensional input space \cite{aggarwal2001surprising}. Hence, we use a deep neural network \cite{goodfellow2016deep}, specifically, a nonlinear encoder to map the input space to a low-dimensional latent space and then apply the SE kernel to the latent space. 
%However, the mapping of high-dimensional data to low-dimensional latent space produced by the the deep neural networks makes it possible to compute a meaningful measure of distance in the latent space \cite{goodfellow2016deep}. 
%In practice,  there could be infinite number of such low-dimensional manifolds. Relying on the superior curve-fitting ability of deep neural net, we could potentially forge a latent space that lies in an Euclidean space while approximating one of the underlying low-dimensional manifold. 

\para{Time-invariant Kernel $k^{(i)}_{\phi}$.}
We introduce a time-invariant kernel to capture cluster correlation, \ie time-invariant correlations among individuals that share similar characteristics. The structure of time-invariant kernel is shown in the bottom part of Figure~\ref{fig:kernel}. Let $\iota(\bx_{i \cdot})=i$ be a mapping function that identifies the individuals, and $g_{\phi}:\iota(\mathcal{X})\rightarrow \mathcal{S}^{(i)}\in\mathbb{R}^{D_i}$ be an embedding function that maps each individual to a vector in the latent space. Then for any pair of data points $\bx_{i\cdot},\bx_{j\cdot}$ with arbitrary observation indices, the time-invariant kernel is given by:
\begin{linenomath}
\begin{equation}
    \kphi(\bx_{i\cdot},\bx_{j\cdot})=k_{SE}(g_{\phi}\circ\iota(\bx_{i\cdot}),g_{\phi}\circ\iota(\bx_{j\cdot}))
\end{equation}
\end{linenomath}

\subsection{Learning a L-DKGPR model from data}
\label{sec:vi}
We now proceed to describe how to efficiently learn an L-DKGPR model and use it to make predictions. Because of space constraints, the details of the derivations are relegated to Appendix A.

\para{Model Inference.}  Our approach to efficiently learning an L-DKGPR model draws inspiration from \cite{wilson2016stochastic}, to greatly simplify the computation of the GP posterior by reducing the effective number of rows in $X$, from $N$ to $M$ ($M\ll N$), where $M$ is the number of  {\em inducing points}. However, unlike \cite{wilson2016stochastic}, which uses inducing points in the input space,  we use  inducing points from {\em a low-dimensional latent space}. Let $Z=\{\bz_m\}_{m=1}^M$ be the collection of inducing points, and $\tu$ their corresponding signal. The kernel computations based on the inducing points are given by:
\begin{linenomath}
\begin{align}
\nonumber
    &\kgamma(\bx,\bz)=k_{SE}(e_\gamma(\bx),\bz) \\ \nonumber
    &\kgamma(\bz_i, \bz_j)=k_{SE}(\bz_i,\bz_j)
\end{align}
\end{linenomath}
Replacing inducing points in the input space with those in a low-dimensional latent space offers several advantages. First, we no longer need to use the encoder network $e_\gamma(\cdot)$ to transform the inducing points $\bz$, thus increasing the computational efficiency of the model. Second, the latent space is dense, continuous, and usually is of much lower dimension than the input space ($D_v \ll P$). The resulting  parameterization of inducing points directly in the latent space, results in a reduction in the number of parameters that describe the inducing points (\ie $Z$) from $\mathcal{O}(MP)$ to $\mathcal{O}(MD_v)$. Third, the latent space simplifies the optimization of L-DKGPR, especially when the input space is defined by heterogeneous data types subject to domain-specific constraints, because the latent space is always continuous regardless the constraints in the input space. We define $\iota(\bz_m)=I+m$ to distinguish the inducing points from the input data. We can now express the joint signal distribution as follows:
\begin{linenomath}
\begin{equation}
\label{eq:f_given_uxz}
    (\tf,\tu|X,Z){\sim}\mathcal{N}\left(
\begin{bmatrix}
    0 \\
    0
\end{bmatrix},\begin{bmatrix}
    K_{XX} & K_{XZ} \\
    K_{XZ}^\top & K_{ZZ}
\end{bmatrix}
\right)
\end{equation}
\end{linenomath}
Therefore, the signal distribution conditioned on the inducing points is given by:
\begin{linenomath}
\begin{equation}
    (\tf|\tu,X,Z){\sim}\mathcal{N}(K_{XZ}K_{ZZ}^{-1}\tu,K_{XX}-K_{XZ}K_{ZZ}^{-1}K_{XZ}^\top)
\end{equation}
\end{linenomath}
Let $\Theta=\{\alpha^{(v)},\alpha^{(i)},\gamma,\phi,\sigma^2, Z\}$ be the model parameters. We aim to learn the parameters by maximizing the log of marginal likelihood $p(\by|X,Z)$. By assuming a variational posterior over the joint signals $q(\tf,\tu|X,Z)=q(\tu|X,Z)p(\tf|\tu,X,Z)$, we can derive the evidence lower bound (see \eg \cite{wilson2016stochastic}):
\begin{linenomath}
\begin{equation}
\label{eq:elbo}
     \mathcal{L} \triangleq \mathbb{E}_{q(\tf,\tu|X,Z)}[\log p(\by|\tf)]-\text{KL}[q(\tu|X,Z)||p(\tu|Z)]
\end{equation}
\end{linenomath}
We define the proposal posterior  $q(\tu|X,Z)=\mathcal{N}(\bm{\mu}_q,L_qL_q^\top)$. To speed up the computation, we follow the deterministic training conditional (DTC) \cite{seeger2003fast}, an elegant sparse method for accurate computation of the Gaussian process posterior by retaining exact likelihood coupled with an approximate posterior \cite{liu2020gaussian}, rendering $(\tf|\tu,X,Z)$ deterministic during the training phase. Letting $A=K_{XZ}K_{ZZ}^{-1}$ and reparameterizing $\tu=\bmu_q+L_q\bep$ with $\bep{\sim}N(\bm{0},\tI)$, we can rewrite the ELBO in closed form:
\begin{linenomath}
\begin{align}
\label{eq:elbo_ana}
\nonumber
    2\mathcal{L} = &-2N\log\sigma - \sigma^{-2} (\Norm{\by}_2^2-2\by^\top A\bmu_{q} + \Norm{A\bmu_{q}}_2^2 \\ \nonumber
    & + \Norm{A L_q \bm{1}}_2^2) -\log \norm{K_{ZZ}} + 2\log \norm{L_q} + M \\
    & - \text{tr}(K_{ZZ}^{-1}L_q L_q^\top) - \bm{\mu}_q^\top K_{ZZ}^{-1} \bm{\mu}_q
\end{align}
\end{linenomath}
where $\bm{1}$ is a column vector of ones. We can then compute the partial derivatives of $\mathcal{L}$ w.r.t. the parameters of the proposal posterior $q(\tu|X,Z)$ (\ie $\{\bmu_q, L_q\}$), yielding:
\begin{linenomath}
\begin{align}
    & \frac{\partial \mathcal{L}}{\partial \bmu_q}=\frac{1}{\sigma^2}(-A^\top \by + A^\top A \bmu_q) + K_{ZZ}^{-1}\bmu_q = 0 \\
    & \frac{\partial \mathcal{L}}{\partial L_q} = \frac{1}{\sigma^2}A^\top A L_q \bm{1}\bm{1}^\top + (L_q^{-\top} + K_{ZZ}^{-1}L_q) = 0
\end{align}
\end{linenomath}
Solving the above equations gives:
\begin{linenomath}
\begin{align}
    & \bmu_q = \sigma^{-2}K_{ZZ}BK_{XZ}^\top \by \label{appeq:muq}\\
    & L_q(\tI + \bm{1}\bm{1}^\top) = K_{ZZ}BK_{ZZ} \label{appeq:lq}
\end{align}
\end{linenomath}
with $B=(K_{ZZ} + \sigma^{-2}K_{XZ}^\top K_{XZ})^{-1}$. To solve the triangular matrix $L_q$ from \eqref{appeq:lq}, we first compute the Cholesky decomposition of $\tI+\bm{1}\bm{1}^\top = CC^\top$ and $K_{ZZ}BK_{ZZ} = UU^T$. We then simplify both side of \eqref{appeq:lq} to $L_qC = U$. $L_q$ can then be solved by exploiting the triangular structure on both side with
\begin{linenomath}
\begin{equation}
\label{appeq:lq_solve}
    L_{i,i-k}= \frac{U_{i,i-k}-\sum_{j=0}^{k-1}L_{i,i-j}C_{i-j,i-k}}{C_{i-k,i-k}}
\end{equation}
\end{linenomath}
where $k=0,\cdots,i-1$, $L_{i,j}$ is a shorthand for $[L_q]_{i,j}$. We separate the model parameters into two groups, \ie parameters w.r.t. the proposal posterior $\{\bmu_q,L_q\}$ and the remaining parameters $\Theta$, and use an EM-like algorithm to update both groups alternatively. The L-DKGPR algorithm is listed in Algorithm~\ref{alg1}. 

\begin{algorithm}[tb]
\DontPrintSemicolon
  \caption{L-DKGPR}
  \label{alg1}
  \KwInput{Training set $S=\left\{X,\by \right\}$, latent dimension $D_v,D_i$, number of inducing points $M$, gradient-based optimizer and its related hyper-parameters (\ie learning rate, weight decay, mini-batch size), alternating frequency $T$.}
  Initialize the parameters $\Theta=\{\sigma^2, Z,\alpha^{(v)},\alpha^{(i)}, \gamma, \phi$\}  \\
  \While{Not converged}
   {
   		Update proposal posterior $q(\tu|X,Z)$ according to \eqref{appeq:muq} and \eqref{appeq:lq_solve} \\
   		$t=0$ \\
   		\For{$t<T$}    
        { 
        	Update $\Theta$ using the input optimizer.\\
        	$t=t+1$
        }
   }
\end{algorithm}

\para{Prediction.}
% A common approximation assumption associated with the inducing points idea is that the signals between training data and test data are conditionally independent given $\tu$ \cite{quinonero2005unifying}. This is particularly useful during the test phase. 
Given the covariate matrix $\Xstar$ for the test data, the predictive distribution is given by:
\begin{linenomath}
\begin{align}
\label{eq:predictive_londkgp} \nonumber
    p(\tf_*|\Xstar,X,& y,Z) \simeq \mathcal{N}(K_{\Xstar Z}(K_{ZZ}+\sigma^2 \tI)^{-1}\bm{\mu}_q,\\
    & K_{\Xstar \Xstar} - K_{\Xstar Z}(K_{ZZ}+\sigma^2 \tI)^{-1}K_{\Xstar Z}^\top)
\end{align}
\end{linenomath}
\para{Complexity.} The time complexity and space complexity of both inference and prediction are $\mathcal{O}(NM^2)$ and $\mathcal{O}(NM)$ respectively, where  $N$ is the number of measured outcomes, and $M$ the number of inducing points.

%% file: experiment.tex
\section{Experiments}
We compare L-DKGPR to several state-of-the-art LDA and GP methods on simulated as well as  real-world benchmark data. The experiments are designed to answer research questions about accuracy, scalability, and interpretability of L-DKGPR: (RQ1) How does the performance of L-DKGPR compare with the state-of-the-art methods on standard longitudinal regression tasks? (RQ2) How does the scalability of L-DKGPR compare with that of the state-of-the-art longitudinal regression models? (RQ3) Can L-DKGPR reliably recover the rich correlation structure from the data? (RQ4) How do the different components of L-DKGPR contribute to its overall performance? (RQ5) What is the advantage of solving the exact ELBO in \eqref{eq:elbo_ana} compared to solving its original form in \eqref{eq:elbo} using Monte Carlo sampling \cite{wilson2016stochastic}?

\subsection{Data}
We used one simulated data set and three real-world longitudinal data sets in our experiments:\footnote{Details of  generation of simulated data  and of pre-processing of real-world data are provided in the Appendix C.}

\noindent\textbf{Simulated data.} We construct simulated longitudinal data that exhibit \ie longitudinal correlation (LC) and multilevel correlation (MC) as follows: The outcome is generated using $\by=f(X)+\bm{\epsilon}$ where $f(X)$ is a non-linear transformation based on the observed covariate matrix $X$ and the residual $\bm{\epsilon}{\sim}N(\bm{0},\Sigma)$. To simulate longitudinal correlation, we simply set $\Sigma$ to a block diagonal matrix with non-zero entries for within-individual observations. To simulate multilevel correlation, we first split the individuals into $C$ clusters and assign non-zero entries for the data points in the same cluster. Following  \cite{cheng2019additive,timonen2019interpretable}, we simulate $40$ individuals, $20$ observations, and $30$ covariates for each individual. We vary the number of clusters $C$ from $[2,5]$.
    
\noindent \textbf{Study of Women’s Health Across the Nation (SWAN)} \cite{sutton2005sex}. SWAN is a multi-site longitudinal study designed to examine the health of women during the midlife years. We consider the task of predicting the CESD score, which is used for screening for depression. Similar to \cite{liang2019lmlfm}, we define the adjusted CESD score by $y=\text{CESD}-15$, thus $y\geq 0$ indicates depression. The variables of interest include aspects of physical and mental health, and demographic factors such as race and income. The resulting data set has $3,300$ individuals, $137$ variables and $28,405$ records. 

\noindent\textbf{General Social Survey (GSS)} \cite{smith2017general}. The GSS data were gathered over 30 years on contemporary American society collected with the goal of understanding and explaining trends and constants in attitudes, behaviors, and attributes. In our experiment, we consider the task of predicting the self-reported general happiness of $4,510$ individuals using $1,553$ features and $59,599$ records. We follow the experimental setup in \cite{liang2019lmlfm}, with $y=1$ indicates happy and $y=-1$ indicates the opposite.
    
\noindent\textbf{The Alzheimer's Disease Prediction Of Longitudinal Evolution (TADPOLE)} \cite{marinescu2018tadpole}. The TADPOLE challenge involves predicting the symptoms related to Alzheimer's Disease (AD) within 1-5 years of a group of high-risk subjects. In our experiment, we focus on predicting the ADAS-Cog13 score using the demographic features and MRI measures (Hippocampus, Fusiform, WholeBrain, Entorhinal, and MidTemp). The resulting data set has $1,681$ individuals, $24$ variables and $8,771$ records. 

\subsection{Experimental Setup}
To answer RQ1, we use both simulated data and real-world data. To evaluate the regression performance, similar to \cite{liang2019lmlfm}, we compute the mean and standard deviation of $\text{R}^2$ between the actual and predicted outcomes of each method on each data set across 10 independent runs. We use 50\%, 20\%, 30\% of data for training, validation, and testing respectively. 

To answer RQ2, we take data from a subset consisting of $50$ individuals with the largest number of observations from each real-world data. We record the run time per iteration of each method on both the $50$-individual subset and full data set. Because not all baseline methods implement GPU acceleration, we compare  the run times of all the methods without GPU acceleration. We report execution failure if a method fails to converge within 48 hours or generates an execution error \cite{liang2019lmlfm}. 

To answer RQ3, we rely mainly on the simulated data since the actual correlation structures underlying the real-world data sets are not known. We evaluate the performance of each method by visualizing the learned correlation matrix and compare it to the ground truth correlation matrix on simulated data. 
Additionally, we illustrate how the correlation matrix learned by  L-DKGPR can provide gain useful insights using a case study with the SWAN data. Results for case study is presented in Appendix D.

To answer RQ4, we compare the performance of L-DKGPR with L-RBF-GPR, a variant that replaces the learned deep kernel with a simple RBF kernel; and L-DKGPR-, a variant of L-DKGPR without the time-invariant effects. 
% We do not compare L-DKGPR with a variant without the time-varying effects since it gives the same predictions for the same individual regardless of time, which is considered unrealistic in LDA applications.

To answer RQ5, we compare the regression performance and hyper-parameter choices of L-DKGPR solved using Algorithm~\ref{alg1} with the version of L-DKGPR solved using Monte Carlo sampling \cite{wilson2016stochastic} on SWAN and GSS data sets.

\vspace*{0.1in}
\noindent\textbf{Baseline Methods}
We compare L-DKGPR with the following baseline methods: (i) Conventional longitudinal regression models, \ie \textbf{GLMM} \cite{bates2014fitting} and \textbf{GEE} \cite{inan2017pgee}; (ii) State-of-the-art longitudinal regression models, \ie \textbf{LMLFM} \cite{liang2019lmlfm} and \textbf{LGPR} \cite{timonen2019interpretable}; (iii) State-of-the-art Gaussian Process models for general regression, \ie \textbf{KISSGP} with deep kernel \cite{wilson2016stochastic} (we use the same deep structure as in our time-varying kernel) and \textbf{ODVGP} \cite{salimbeni2018orthogonally}. Implementation details\footnote{Data, codes and Appendix are available at \url{https://github.com/junjieliang672/L-DKGPR}.} and hyper-parameter settings of L-DKGPR as well as the baseline approaches are provided in Appendix B.

\subsection{Results}
We report the results of our experiments designed to answer the research questions RQ1-RQ4.

\begin{table*}[tb]
% \vspace{-0.02cm}
\small
\caption{Regression accuracy R$^2$ (\%) comparison on simulated data with different correlation structures.}
\begin{center}
% \scalebox{0.7}{
\begin{tabular}{c|ccccc}
\toprule
Method & LC & MC($C=2$) & MC($C=3$) & MC($C=4$) & MC($C=5$) \\
\midrule
L-DKGPR                 & \textbf{86.0$\pm$0.2}    & \textbf{91.3$\pm$0.2}           & \textbf{99.6$\pm$0.2}           & \textbf{99.8$\pm$0.2}           & \textbf{99.8$\pm$0.2}                                                                                   \\
KISSGP &  85.9$\pm$1.7    & -43.4$\pm$33.3  &  -55.5$\pm$7.1 &  -58.2$\pm$14.4 & -57.2$\pm$17.9 \\
ODVGP  &  82.3$\pm$5.2  &  -1.6$\pm$16.9  &  -14.7$\pm$6.5 & -13.5$\pm$8.4  & -6.1$\pm$4.4 \\
LGPR                    & -37.1$\pm$19.1   & -123.6$\pm$162.0           & -26.3$\pm$43.2           & -9.1$\pm$14.8           & -0.1$\pm$5.9           \\
LMLFM                   & 54.7$\pm$15.1   & -138.3$\pm$121.9           & -48.3$\pm$123.6           & 22.6$\pm$49.0           & 36.2$\pm$41.1                                                                                    \\
GLMM                    & 5.3$\pm$27.9   & -656.3$\pm$719.8           & -801.4$\pm$507.4           & -684.1$\pm$491.3           & -528.7$\pm$313.5                                                                                   \\
GEE                     & 59.0$\pm$24.5   & -636.1$\pm$606.0           & -703.6$\pm$465.8           & -665.6$\pm$554.3           & -516.5$\pm$457.5           \\
\bottomrule
\end{tabular}
% }
\end{center}
\label{tab:simulation}
% \vspace{-0.06cm}
\end{table*}

\begin{table*}[tb]
% \vspace{-0.02cm}
\small
\caption{Regression accuracy R$^2$ (\%) on real-world data sets. We use `N/A' to denote execution error.}
\begin{center}
% \scalebox{0.8}{
\begin{tabular}{cccc|ccccccc}
\toprule
Data sets & $N$ & $I$ & $P$ & L-DKGPR  &  KISSGP  &  ODVGP  & LGPR           & LMLFM        & GLMM         & GEE   \\
\midrule
TADPOLE                    & 595   & 50                   & 24  & 44.0$\pm$5.6 & 1.2$\pm$10.1 & 9.0$\pm$14.1  & -261.1$\pm$9.0 & 8.7$\pm$5.1  & \textbf{50.8$\pm$5.5} & -11.4$\pm$4.8   \\
SWAN                       & 550    & 50                   & 137                      & \textbf{46.8$\pm$4.9} & 42.4$\pm$4.6 & 29.0$\pm$3.1 & -16.6$\pm$12.7 & 38.6$\pm$4.2 &   40.1$\pm$7.7           &    46.4$\pm$8.0   \\
GSS                        & 1,500  & 50                   & 1,553  & \textbf{19.1$\pm$3.7} & 12.5$\pm$6.3 & -7.6$\pm$3.3 & N/A            & 15.3$\pm$1.4 &    N/A          & -4.6$\pm$3.5 \\ \hline

TADPOLE                    & 8,771  & 1,681                & 24               & \textbf{64.9$\pm$1.4} & 0.6$\pm$3.9 & 21.1$\pm$1.0 & N/A            & 10.4$\pm$0.6 & 61.9$\pm$1.9 & 17.6$\pm$0.7  \\
SWAN                       & 28,405 & 3,300                & 137               & \textbf{52.5$\pm$0.4} & 20.5$\pm$7.6  & 24.9$\pm$21.8  & N/A            & 48.6$\pm$2.0 & N/A          & N/A          \\
GSS                        & 59,599 & 4,510                & 1,553               & \textbf{56.9$\pm$0.1} & 53.1$\pm$0.9 & 15.4$\pm$27.0   & N/A            & 54.8$\pm$2.2 & N/A          & N/A    \\
\bottomrule
\end{tabular}
% }
\end{center}
\label{tab:real_life}
% \vspace{-0.02cm}
\end{table*}

\noindent\textbf{L-DKGPR vs baseline longitudinal regression methods.}
The results are reported in Table~\ref{tab:simulation} and Table~\ref{tab:real_life} for simulated and real-world data sets respectively. In the case of simulated data, we find that KISSGP, ODVGP, GEE and GLMM fail in the presence of MC with the mean $\text{R}^2$ being negative (indicative of models containing variables that are not predictive of the response variable). This can be explained by the fact that GEE is designed only to handle pure LC, thus fails to account for CC or MC. While GLMM is capable of handling MC, it requires practitioners to specify the cluster structure responsible for CC prior to model fitting. However, in our experiments, the cluster structure is unknown a priori. Hence it is not surprising that  GLMM performs poorly. Though both KISSGP and ODVGP are conceptually viable to handle data with complex correlation, they both experience dramatic performance drop when cluster correlation (or time-invariant effects) are presented. Moreover, we find that although LMLFM outperforms GLMM and GEE in the presence of MC, its $\text{R}^2$ is still quite low. This is because LMLFM accounts for only a special case of MC, namely, for CC among individuals observed at the same time points, and not all of the CC present in the data. We find that LGPR performs rather poorly on both simulated and real-world data. This might due to the fact that LGPR obtains the contributions of each variable to the kernel independently before calculating their weighted sum. Though it is possible to incorporate higher-order interactions between variables into LGPR, doing so requires estimating large numbers of interaction parameters, with its attendant challenges, especially when working with small populations. In contrast to the baseline methods, L-DKGPR consistently and significantly outperforms the baselines by a large margin. On the real-world data sets, L-DKGPR outperforms the longitudinal baselines in most of the cases. 
%The superior performance of L-DKGPR can perhaps be explained by two factors. First, because L-DKGPR uses a learned deep kernel, it is capable of accommodating complex correlation among the data points. Second, the proposed kernel structure in L-DKGPR is particularly suitable for handling longitudinal data with both time-varying and time-invariant effects. 

\noindent\textbf{Scalability of L-DKGPR vs. baseline methods.}
We see from Table~\ref{tab:real_life} that most longitudinal baselines, \ie LGPR, GLMM, and GEE, fail to process real-world data sets with large numbers of covariates. Indeed, their computational complexity increases in proportion to $P^3$ where $P$ is the number of covariates. In contrast, L-DKGPR, LMLFM and state-of-the-art GP baselines  (KISSGP and ODVGP) scale gracefully with increasing number of data points and covariates. For  CPU run time analysis, please refer to Appendix D.

\begin{figure}[t!]
% \vspace{-0.2cm}
\centering 
\includegraphics[width=1.05\linewidth]{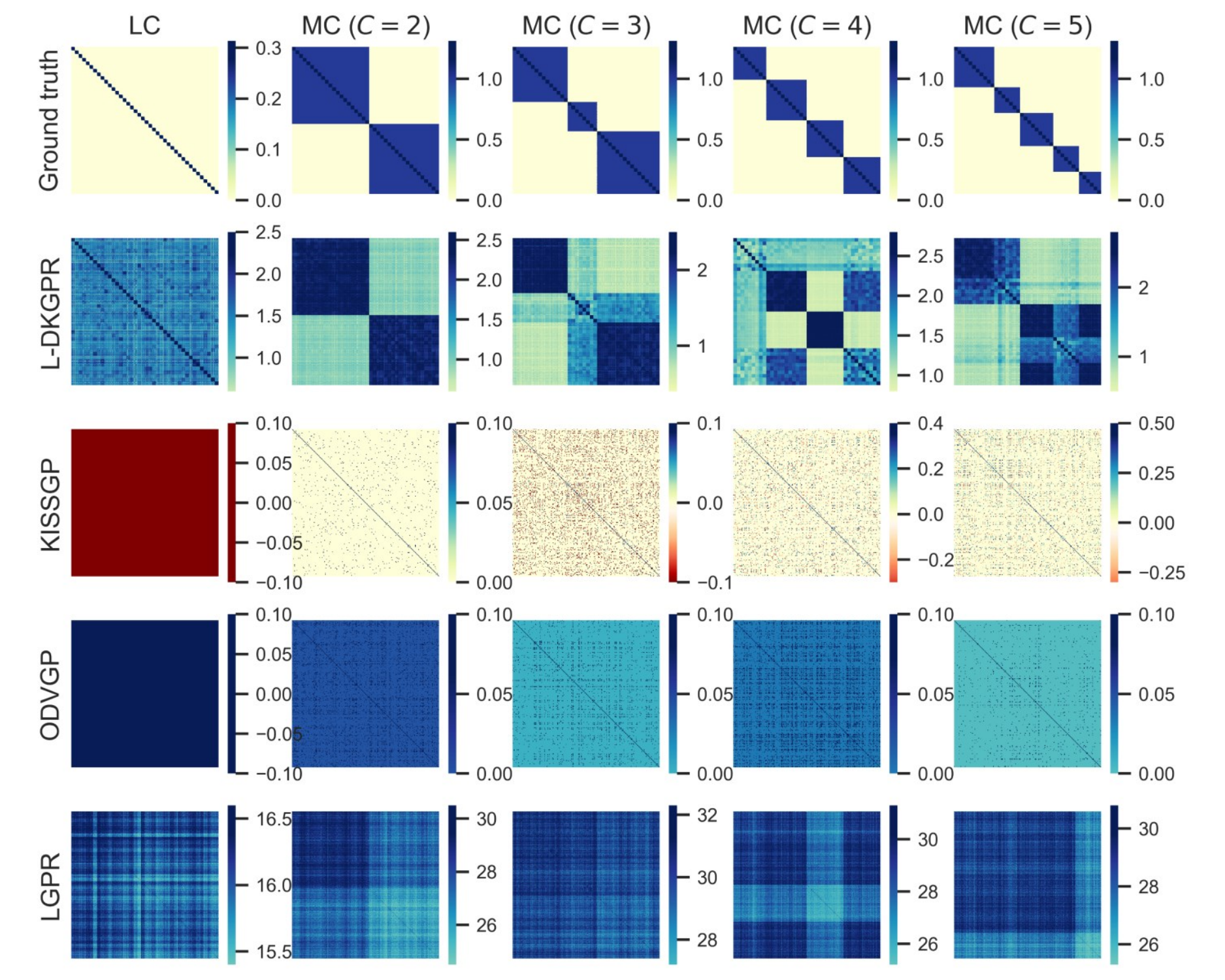}
\caption{Outcome correlation estimated by all GP methods on simulated data.}
\label{fig:corr_compare}
% \vspace{-0.12cm}
\end{figure}

\noindent\textbf{Recovery of Correlation Structure.}
The outcome correlations estimated by all GP methods on the simulated data are shown in Figure~\ref{fig:corr_compare}. We see that KISSGP and ODVGP are incapable of recovering any correlation structure from the data. LGPR seems to be slightly better than KISSGP and ODVGP when MC is presented. However, we see that only one known cluster is correctly recovered when $C>2$. This suggests that these methods fail to recover accurate correlation structures, which is consistent with their poor performance in terms of $\text{R}^2$. In contrast, L-DKGPR is able to recover most of the correlation structure present in the data. It is worth noting that recovering correlation structure is a challenging task and although L-DKGPR is the best performing model, the learned correlation structure is still far from perfect. A possible explanation is that without further prior constraints on the kernel structure, the kernel search space is very large. Since L-DKGPR works in an MLE framework, it searches for a kernel to improve the likelihood. When optimal solution is surrounded by infinitely many local maxima, each with simpler kernel structure but sufficiently high likelihood, it is not surprising that L-DKGPR gets stuck in one of such local maxima since the kernel initialization of L-DKGPR is uninformative.

\begin{table}[t]
% \vspace{-0.12cm}
\small
\caption{Effect on the regression accuracy R$^2$ (\%) of different components of L-DKGPR}
\begin{center}
\scalebox{0.9}{
\begin{tabular}{c|cccc}
\toprule
Data sets  & L-DKGPR                                & L-DKGPR-v    & L-DKGPR-i      & L-RBF-GPR \\
\midrule
TADPOLE                   & \textbf{64.9$\pm$1.4} & 13.2$\pm$1.1 & 56.3$\pm$1.3 & 55.5$\pm$2.4      \\
SWAN                      & \textbf{52.5$\pm$0.4} & 29.0$\pm$3.2 & 16.7$\pm$ 2.4 &  5.4$\pm$1.6     \\
GSS                       & \textbf{56.9$\pm$0.1} & 56.2$\pm$0.1 & -0.2$\pm$0.2 & -14.1$\pm$0.4     \\
\bottomrule
\end{tabular}
}
\end{center}
\label{tab:ablation_r2}
% \vspace{-0.2cm}
\end{table}

\noindent\textbf{Ablation study.}
Regression accuracy comparison on complete real-world data sets is shown in Table~\ref{tab:ablation_r2}. \textbf{Role of time-invariant component}: We see a dramatic drop in regression performance when time-invariant effects are not modeled (L-DKGPR-v) as compared to when they are (L-DKGPR). This result underscores the importance of modeling the time-independent components of LC and CC for accurate modeling of longitudinal data. This task is simplified by the decomposition of the correlation structure into the time-varying and time-invariant components. The time-invariant component is analogous to estimating the mean correlation whereas the time-varying component contributes to the residual. Hence, the decomposition of the correlation structure into time-varying and time-invariant components should help reduce the variance of the correlation estimates. \textbf{Role of time-varying component}: We observe significant performance drop when time-varying effects are not modeled (L-DKGPR-i) as compared to L-DKGPR. This is reasonable because without the time-varying kernel, the model gives the same outcome prediction for an individual at all time. This is unrealistic for longitudinal data. \textbf{Role of deep kernel}: L-DKGPR consistently outperforms L-RBF-GPR (which uses RBF kernel instead of the deep kernel used by L-DKGPR), with the performance gap between between the two increasing with increase in the number of covariates. This is perhaps explained by the pitfalls of Euclidean distance as a measure of similarity between data points in a high dimensional data space \cite{aggarwal2001surprising} (and hence  kernels such as the RBF kernel which rely on Euclidian distance in the data space), and the apparent ability of the learned deep kernel to perform such similarity computations in a low-dimensional latent space where the computed similarities are far more reliable. 

\noindent\textbf{Effect of solving the exact ELBO with Algorithm~\ref{alg1}.}
Table~\ref{tab:effect_elbo} presents the results in comparing L-DKGPR solved using Algorithm~\ref{alg1} with a version of L-DKGPR solved using the vanilla Monte Carlo sampling \cite{wilson2016stochastic}. We find that under the same hyper-parameter setting, our solver outperforms the sampling solver by a large margin. To ensure similar regression performance, we have to modify the hyper-parameters for the sampling solver by increasing the number of inducing points $M$ to $128$ and using about $10$ times more training iterations. The result indicates that coping with the variance of the noisy ELBO approximation increases the number of parameters and hence the number of iterations needed. 

\begin{table}[t]
\small
% \vspace{-0.12cm}
\caption{Effect of solving L-DKGPR using Algorithm~\ref{alg1} vs. Monte Carlo sampling.}
\begin{center}
% \scalebox{0.9}{
\begin{tabular}{c|cccc}
\toprule
Data sets             & Solver                                     & $M$ & Iterations & R$^2$ (\%) \\
\midrule
\multirow{3}{*}{SWAN} & Alg.~\ref{alg1} &  10   &  300          &      \textbf{52.5$\pm$0.4}      \\
                      & Sampling                               &    10 &      300      &     3.1$\pm$0.2       \\
                      & Sampling                                &     128&  3,000          &       51.4$\pm$0.4     \\
\midrule
\multirow{3}{*}{GSS}  & Alg.~\ref{alg1} &   10  & 300           &  \textbf{56.9$\pm$0.1}          \\
                      & Sampling                               &   10  &    300        &4.5$\pm$0.1           \\
                      & Sampling                               &   128  &  3,000          &      55.6$\pm$0.1     \\
\bottomrule
\end{tabular}
% }
\end{center}
\label{tab:effect_elbo}
% \vspace{-0.2cm}
\end{table}

\begin{figure}[t!]
% \vspace{-0.12cm}
\centering 
\includegraphics[width=.7\linewidth]{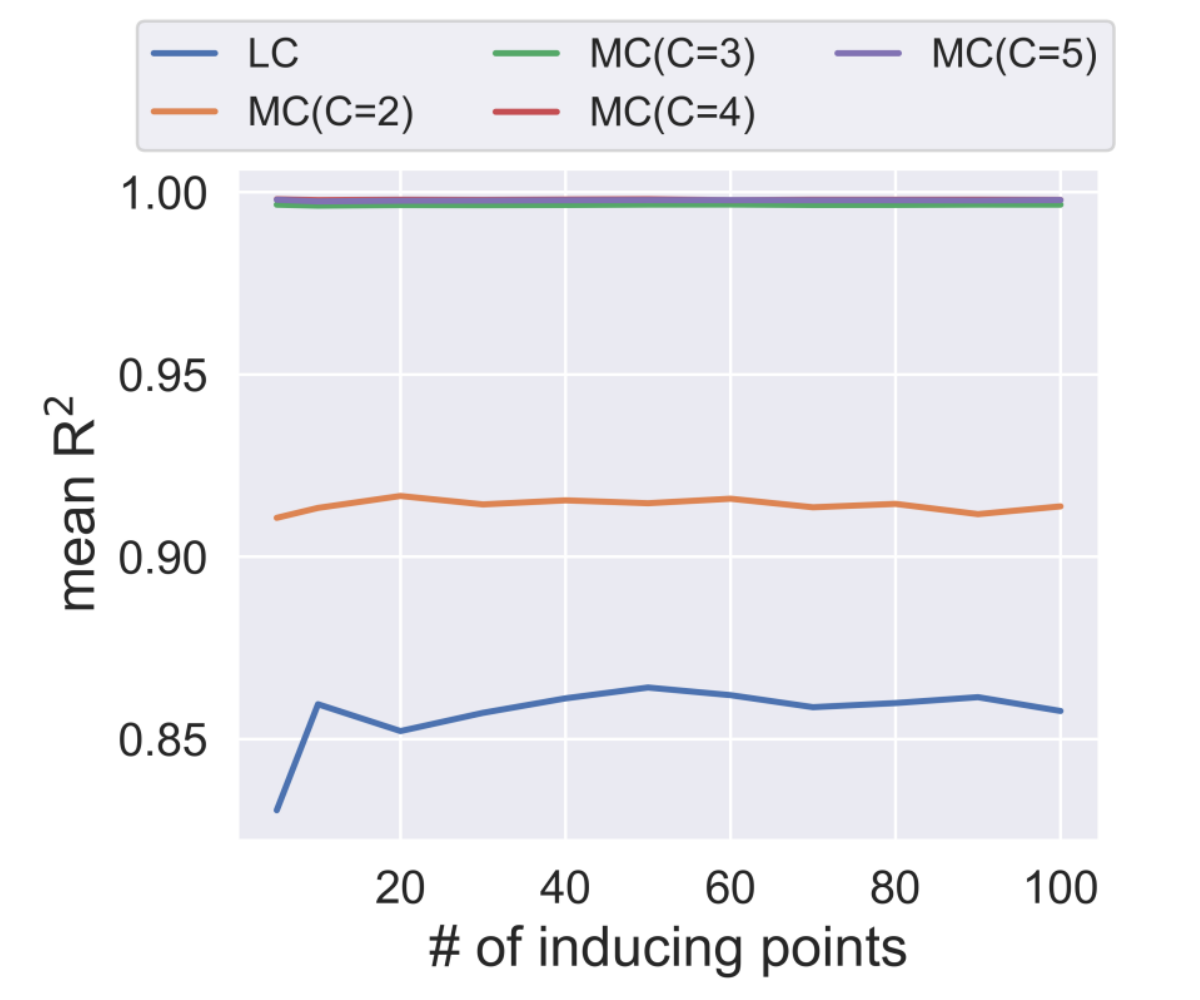}
\caption{Regression performance with different numbers of inducing points on simulated data.}
\label{fig:inducing_points}
% \vspace{-0.2cm}
\end{figure}

\noindent\textbf{Effect of the number of inducing points $M$.} 
\label{app:inducing_points}
Inducing points provide a trade-off between approximation accuracy and efficiency in sparse GP methods. In this experiment, we vary the number of inducing points $M$ from $5$ to $100$ on simulated data and record the R$^2$ as shown in Figure~\ref{fig:inducing_points}. We find that when the number of inducing points reaches a certain threshold, \ie $10$ in \textit{all} simulated settings, regression performance is rather stable, an observation that is supported by our experiments with real-world data as well (results omitted). A theoretical study \cite{burt2019rates} points out that when input data are normally distributed and inducing points are drawn from a $k$-deterministic point process with an SE-ARD kernel, then $M=\mathcal{O}(\log^P N)$. In our case, since the inducing points lie in the latent space, the number of inducing points suffice to our simulated data should be as large as $M=[2\log(40)]^{10}$. In contrast, we empirically show that $M\approx \log N$ is sufficient to get consistent and appealing results. We conjecture that this is because instead of drawing the inducing points from a $k$-DPP process from the input data, we optimize representation of the inducing points jointly with the other model parameters, thus delivering more effective inducing points that summarize the variance of the input data. Proving or disproving this conjecture would require a deeper theoretical analysis of L-DKGPR. 

%% file: conclusion.tex
\section{Conclusion}
We have presented L-DKGPR, a novel longitudinal deep kernel Gaussian process regression model that overcomes some of the key limitations of existing state-of-the-art GP regression methods for predictive modeling from longitudinal data. L-DKGPR fully automates the discovery of  complex  multi-level correlations from longitudinal data.  It incorporates a deep kernel learning method that combines the expressive power of deep neural networks with the flexibility of non-parametric kernel methods, to capture the complex multilevel correlation structure from longitudinal data.  L-DKGPR uses a novel additive kernel that simultaneously models both time-varying and the time-invariant effects. We have shown how L-DKGPR can be efficiently trained using latent space inducing points and the stochastic variational method. We report results of extensive experiments using both simulated and real-world benchmark longitudinal data sets that demonstrate the superior predictive accuracy as well as scalability of L-DKGPR over the state-of-the-art LDA and GP methods. A case study with a real-world data set illustrates the potential of L-DKGPR as a source of useful insights from complex longitudinal data.

\newpage
\noindent
{\bf Acknowledgements:}
This work was funded in part by the  NIH NCATS  grant UL1 TR002014 and by NSF  grants 2041759,  1636795, the Edward Frymoyer Endowed Professorship  at Pennsylvania State University and the Sudha Murty Distinguished Visiting Chair in Neurocomputing and Data Science funded by the Pratiksha Trust at the Indian Institute of Science (both held by Vasant Honavar).